\documentclass[journal]{IEEEtran}
\usepackage{subcaption}
\usepackage{graphicx}
\usepackage{booktabs} 
\usepackage{amsmath}
\usepackage{textcomp}
\usepackage{listings}
\usepackage{xcolor}
\usepackage{threeparttable}
\usepackage{multirow}
\usepackage{array,booktabs}
\graphicspath{ {images/}}
\usepackage[hidelinks]{hyperref}
\usepackage{url}
\usepackage{cite}
\usepackage{algorithm}
\usepackage{algpseudocode}

\usepackage[capitalise,nameinlink]{cleveref}

\definecolor{codeblack}{rgb}{0, 0, 0}
\definecolor{codegreen}{rgb}{0.2, 0.5, 0.2}

\lstdefinestyle{mystyle}{
    commentstyle=\itshape\color{codegreen},
    keywordstyle=\color{magenta},
    basicstyle=\ttfamily\footnotesize\color{codeblack},
    breakatwhitespace=false,         
    breaklines=true,                 
    captionpos=b,                    
    keepspaces=true,                 
    numbersep=5pt,                  
}

\ifCLASSINFOpdf
\else
\fi
\begin{document}

\title{When Foundation Model Meets Federated Learning: Motivations, Challenges, and Future Directions}
%
%
%


\author{Weiming~Zhuang, 
        Chen~Chen, 
        Jingtao~Li, 
        Chaochao~Chen,
        Yaochu~Jin,
        and~Lingjuan~Lyu
\thanks{Weiming~Zhuang, Chen~Chen, Jingtao~Li, and Lingjuan~Lyu are with Sony Research. Email: Weiming.Zhuang@sony.com; ChenA.Chen@sony.com; Jingtao.Li@sony.com; Lingjuan.Lv@sony.com.} 
\thanks{Chaochao Chen is with Zhejiang University, China. E-mail: zjuccc@zju.edu.cn.}
\thanks{Yaochu Jin is with Westlake University, China. E-mail: jinyaochu@westlake.edu.cn.}
}

\maketitle

\begin{abstract}

The intersection of Foundation Model (FM) and Federated Learning (FL) 
presents a unique opportunity to unlock new possibilities 
for real-world applications. On the one hand, FL, as a collaborative learning paradigm, help address challenges in FM development by expanding data availability, enabling computation sharing, facilitating the collaborative development of FMs, tackling continuous data update, avoiding FM monopoly, response delay and FM service down. On the other hand, FM, equipped with pre-trained knowledge and exceptional performance, can serve as a robust starting point for FL. It can also generate synthetic data to enrich data diversity and enhance overall performance of FL. Meanwhile, FM unlocks new sharing paradigm and multi-task and multi-modality capabilities for FL. By examining the interplay between FL and FM, this paper presents the motivations, challenges, and future directions of empowering FL with FM and empowering FM with FL. We hope that this work provides a good foundation to inspire future research efforts to drive advancements in both fields.
\end{abstract}

\begin{IEEEkeywords}
Federated learning, Foundation model
\end{IEEEkeywords}

%
\IEEEpeerreviewmaketitle

\section{Introduction}

\begin{figure}[t!]
\centering
    \includegraphics[width=0.48\textwidth]{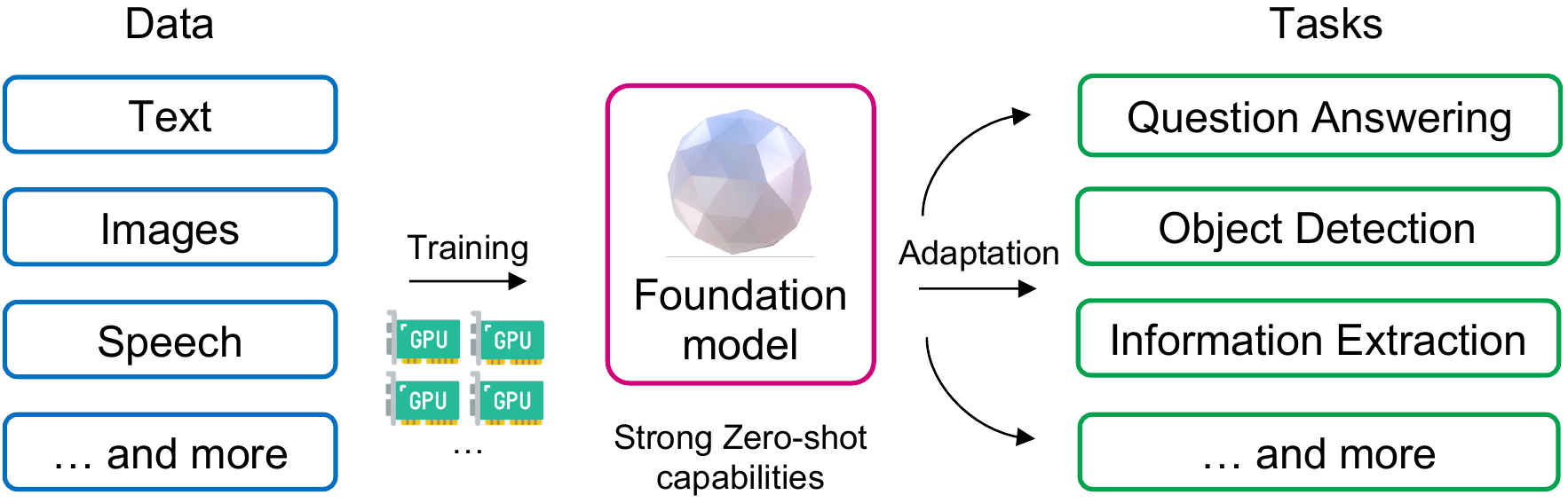}
\caption{Illustration of foundation models.}
\label{fig:intro-fm} 
\end{figure}

\IEEEPARstart{T}{he} field of Artificial Intelligence (AI) has witnessed remarkable advancements in recent years, with the emergence of powerful models and innovative learning techniques. Among them, Foundation Model (FM) represents a significant breakthrough that has revolutionized AI research and applications \cite{bommasani2021fm-survey}. FM has experienced rapid growth, from Large Language Models (LLM) like GPT \cite{zhang2023gpt,radford2019gpt2,brown2020gpt3, openai2023gpt4} and LLaMA \cite{touvron2023llama} to multimodal and vision foundation models \cite{radford2021clip,team2024gemini,xiao2023florence}. These models introduce a new era in AI models characterized by enormous size, pre-trained knowledge, and exceptional performance across a wide range of tasks, as shown in \cref{fig:intro-fm}. FMs have showcased remarkable zero-shot capabilities in understanding natural language and generating coherent responses. However, achieving these capabilities often requires an extensive model size trained with massive datasets using high computational resources.
Further development and utilization of FMs face challenges such as limited availability of large-scale and high-quality public data \cite{villalobos2022will} and high demand on computation resources \cite{bommasani2021fm-survey}.

\begin{figure}[t!]
    \centering
        \includegraphics[width=0.48\textwidth]{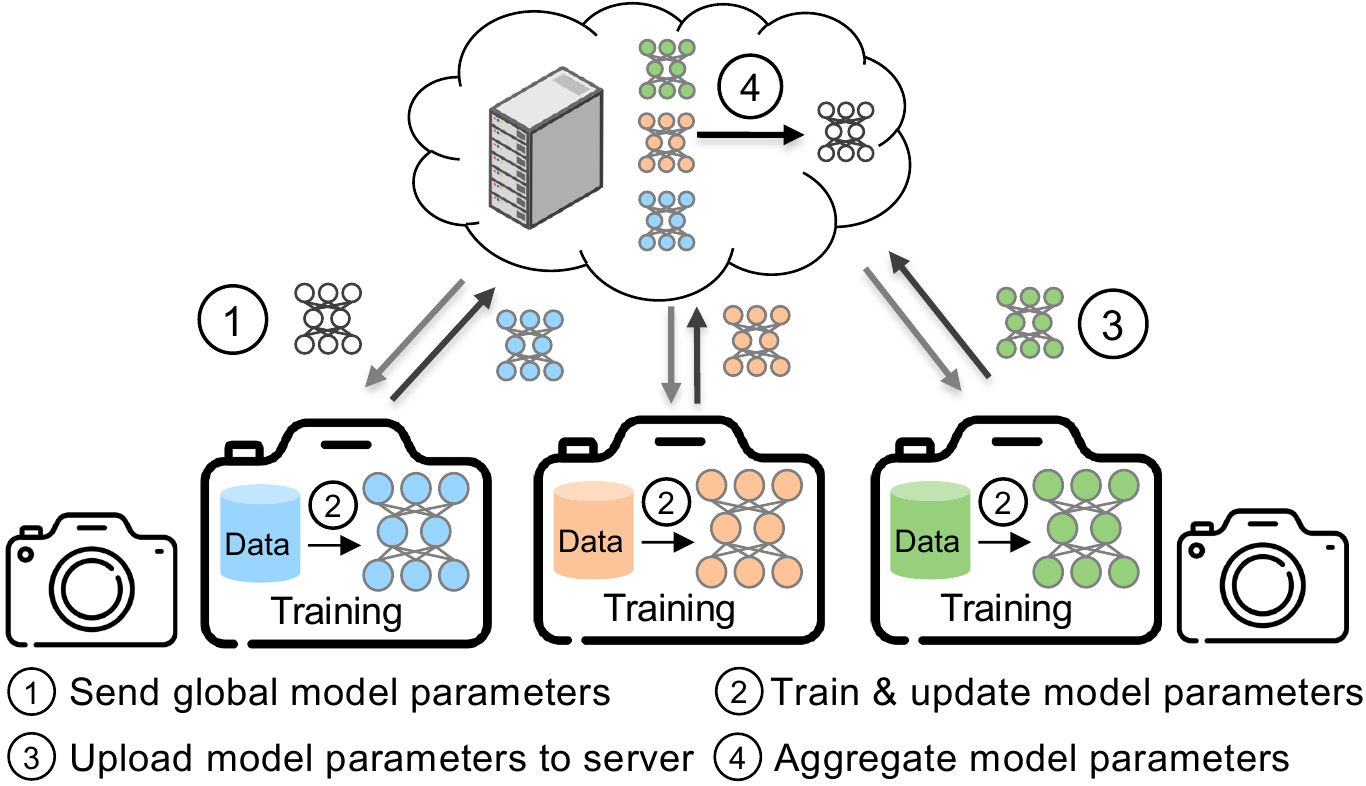}
        \caption{Illustration of federated learning.}
        \vspace{-3ex}
    \label{fig:intro-fl} 
\end{figure}

Federated learning (FL) \cite{fedavg}, on the other hand, is a collaborative learning paradigm that enables in-situ training on distributed data sources. \cref{fig:intro-fl} illustrates the process of FL that enhances data privacy by only transferring model updates from clients to the server instead of raw data. FL has empowered a wide range of applications, including healthcare \cite{li2019brain-tumor1,sheller2018brain-tumor2}, finance \cite{zhang2023privacy,liu2023efficient}, video surveillance \cite{zhuang2020fedreid,zhuang2021fedureid,zhuang2022fedfr,zhuang2022fedreid}, consumer products \cite{hard2018gboard}, and more. FL offers two primary approaches: cross-silo FL and cross-device FL \cite{kairouz2021advances}. Cross-silo FL trains models across multiple institutions, where the number of clients is normally within 100. Cross-device FL trains models across distributed devices, where the number of clients can be up to $10^{10}$. FL supports training a randomly initialized model from scratch or fine-tuning existing models, 
thus empowering scalable and privacy-preserving model development.

\begin{figure*}[t!]
\centering
\includegraphics[width=\textwidth]{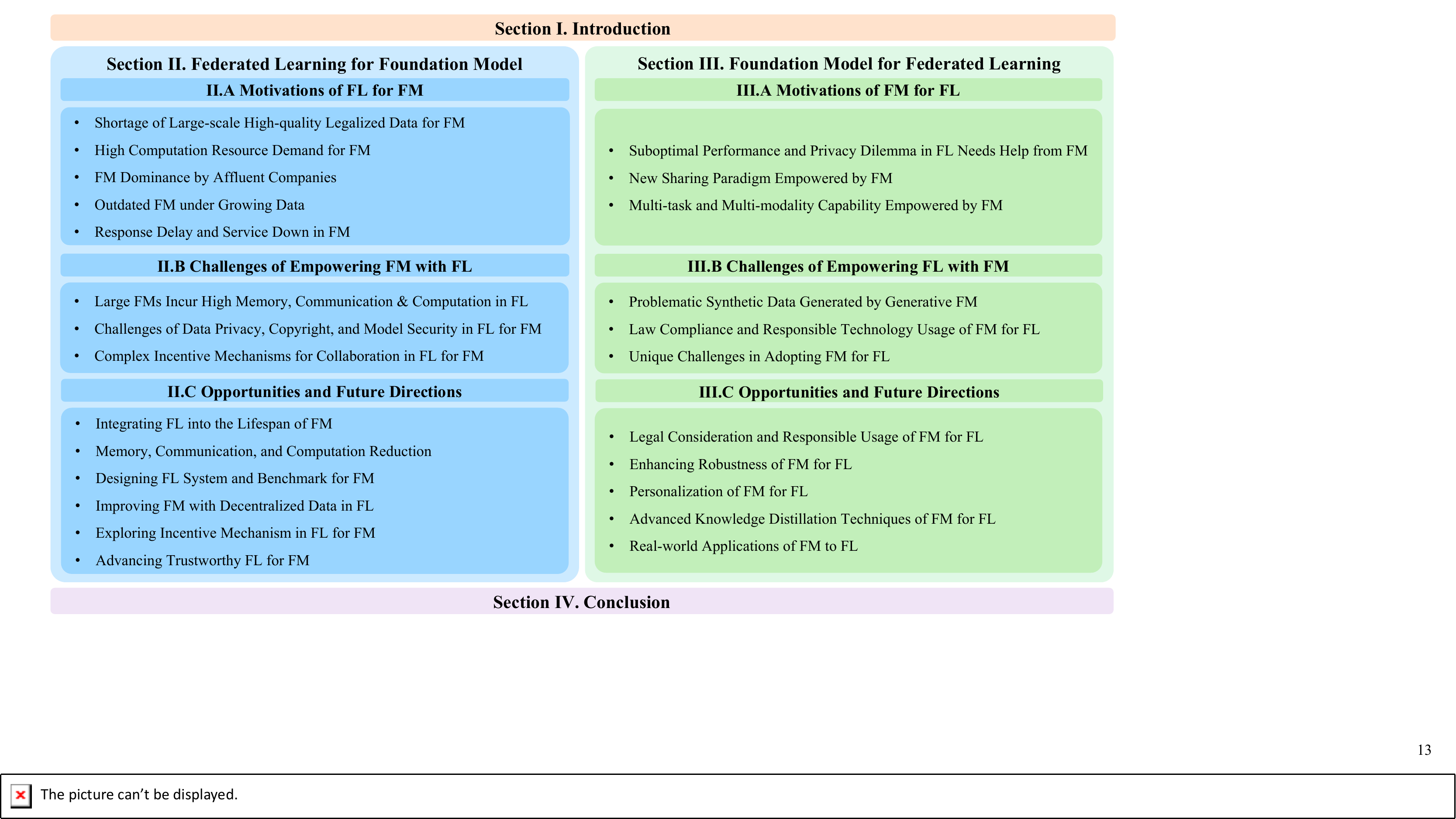}
\caption{Organization of the paper.}
\label{fig:paper-organization} 
\end{figure*}

FL and FM share mutual benefits, each providing unique advantages to the other. On the one hand, FL can play a vital role in mitigating challenges encountered in FM development. For example, FL can expand the availability of data for FM by leveraging distributed data sources. It also enables computation sharing and distributes the training across multiple participants (e.g., devices or organizations). FL also promotes the collaborative development of FMs, democratizing the process and allowing a wider range of participants to contribute. On the other hand, FM can help address challenges in FL. With its pre-trained knowledge and exceptional performance, FM serves as a good starting point, facilitating faster convergence and better performance, especially under non-iid (non-independent and identically distributed) and skewed settings \cite{nguyen2022begin,chen2023importance}. Besides, leveraging FM to generate synthetic data can enrich data diversity, reduce overfitting, and preserve privacy by supplementing or replacing sensitive data with synthetic data. 
Furthermore, multi-modal FMs~\cite{achiam2023gpt, team2024gemini} can be used to annotate unlabeled data, or filter out low-quality data on a wide range of CV and NLP tasks, which helps FL participants to reduce the labeling cost. This mutual reinforcement between FL and FM could create a synergistic relationship that drives advancements in both fields. While FedFM \cite{ren2024advances} explores how FL empowers FM, we provide a comprehensive discussion on their mutual benefits.

This paper aims to explore the motivations, challenges, and future directions of 
the interplay between FL and FM. \cref{fig:paper-organization} summarizes the organization of the paper. 
We explain motivations and delve into the specific challenges faced by FL and FM in their interplay, highlighting their interconnections. 
Furthermore, we examine the opportunities that arise from this intersection. 
By examining these challenges and opportunities, we aim to pave the way for a deeper understanding of the potential of FL in conjunction with the FM and inspire future research directions that can further drive both fields. Throughout this work, we will interchangeably use participants/clients/users/entities to represent the participants in FL.

\section{Federated Learning for Foundation Models}

In this section, we introduce the motivations and challenges faced during the development of Foundation Model (FM) and explain how Federated Learning (FL) serves as a promising solution in opportunities and future directions. 

\subsection{Motivations of FL for FM}

\textbf{Shortage of Large-scale High-quality Legalized Data for FM.}

FM faces a significant challenge in accessing ample high-quality public data for training or fine-tuning \cite{villalobos2022will}. Research predicts the depletion of high-quality language data by 2026, followed by the exhaustion of low-quality language and image data between 2030 and 2050 \cite{villalobos2022will}. As time progresses, the diminishing availability of such data poses a potential obstacle to the continued advancement of FM. While using generated data for training is intuitive, it risks model collapse and the forgetting of underlying data distribution over time \cite{shumailov2023curse}.

Furthermore, publicly available data 
raise a series of concerns. 
Many companies 
scrape data from the Internet for training purposes, leaving individuals uncertain about the usage of their private data, which calls for additional detection of unauthorized data usages~\cite{wang2023diagnosis}. The data scraping and usage practices around large models are still being explored by regulators, and there are lawsuits related to whether scraping publicly available data without compensation or permission constitutes an infringement. Previously free data platforms like Twitter and Reddit have started charging high annual premiums for data access \cite{twitter, reddit}, and 
StackOverflow is contemplating charges \cite{stackoverflow}. Additionally, the LAION-5B dataset, which 
was used by many companies to train models like Stable Diffusion \cite{rombach2022stablediffusion} and ImaGen~\cite{saharia2022imagen} for text-to-image tasks, is entangled in serious copyright issues \cite{chen2023pathway}. Models trained on such problematic datasets may face legal and copyright challenges.

FL emerges as a viable solution that reduces the need for unauthorized scraping of extensive Internet data. It enables FMs to utilize rich data from diverse private sources, enhancing both zero-shot capabilities and adaptability to downstream use cases. Additionally, through the integration of FL with FM, the utilization of private data becomes more accountable, allowing individuals to trace data sources. This enhanced transparency empowers users to selectively unlearn specific data instances from the FM \cite{liu2020federated,che2023fast}. FL 
also reduces labor costs associated with centralized data collection, as data remains localized with participants.

\textbf{High Computational Resource Demand for FM}.
Training large-scale FMs demands substantial computational resources. For instance, LLaMA 3, particularly the larger 405B parameter model \cite{dubey2024llama} used more than 16,000 Nvidia's H100 GPUs, while GPT-4~\cite{achiam2023gpt} and Gemini Ultra~\cite{team2024gemini} have an estimated pre-training hardware utilization cost of \$78 million and \$191 million respectively according to Stanford's 2024 AI Index Report~\cite{maslej2024ai}. Even the vision transformer \cite{dosovitskiy2020image} (ViT/L) employed in CLIP \cite{radford2021clip} requires 8 core TPUv3 for approximately 30 days of training. This poses a challenge for individuals or small organizations that lack the infrastructure or capacity to train FMs independently. 

\begin{figure}[t!]
\centering
\begin{subfigure}[t]{0.22\textwidth}
    \includegraphics[width=\textwidth]{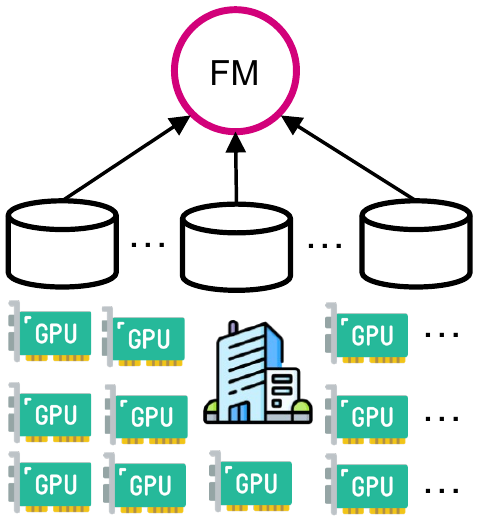}
    \caption{FM trained by one entity}
\end{subfigure}
\hfill
\begin{subfigure}[t]{0.25\textwidth}
    \includegraphics[width=\textwidth]{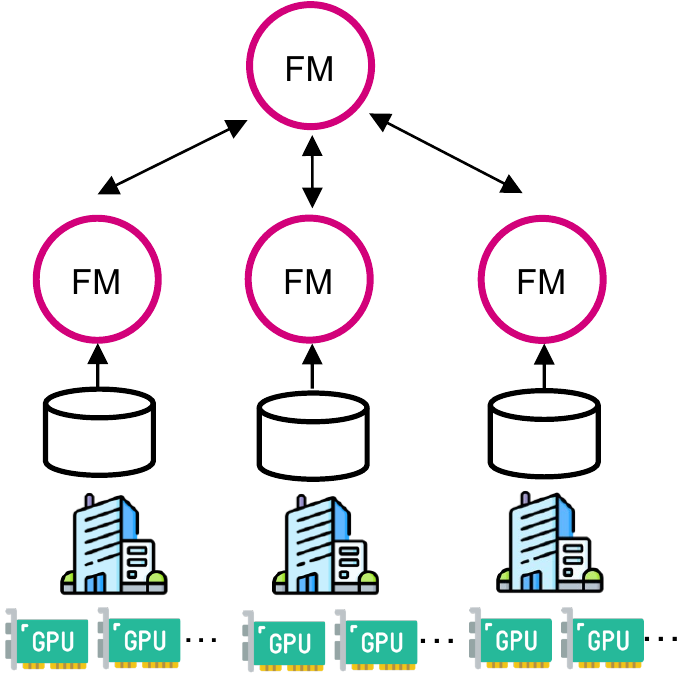}
    \caption{FM trained by multiple entities in FL}
\end{subfigure}
\caption{Resource sharing empowered by FL for FM.}
\label{fig:fl-for-fm-computation} 
\end{figure}

FL mitigates this problem 
via computation sharing. As illustrated in Figure~\ref{fig:fl-for-fm-computation}, 
FL allows participants to pool their computational power, thereby distributing the training burden among participants. 
This resource-sharing approach promotes the development and evolution of the FM by making it accessible to a wider range of participants.

\textbf{FM Dominance by Affluent Companies.}
The development of FM has primarily been driven by industry giants, well-funded startups, and government-supported companies, which limits the participation of smaller organizations. 
While some open-source FM projects exist (e.g.,Vicuna \cite{chiang2023vicuna}), they often rely on FMs open-sourced from large companies like Meta's LLaMA \cite{touvron2023llama, dubey2024llama}, Alibaba's Qwen 2~\cite{wang2024qwen2}, and Stability AI's Stable Diffusion~\cite{rombach2022high}. In contrast, OpenAI's GPT-4 \cite{achiam2023gpt} and Google's Gemini \cite{team2024gemini} remain closed-source.
The expertise, resources, and infrastructure required for advancing FMs are predominantly concentrated within few companies, creating barriers to broader FM development and impeding collective progress in the field. 

FL can dismantle the exclusivity barriers and facilitate collaborative FM development from a wider range of organizations. It can democratize the FM development process by pooling resources from diverse participants. 
This shift towards collaboration promotes that FM development is not confined to a few privileged companies but a collective effort.

\textbf{Outdated FM under Growing Data.}
Real-world scenarios involve continuously updating data, such as data collected by edge devices and IoT sensors \cite{pellegrini2021continual,aggarwal2023chameleon}. However, 
most FMs are developed based on currently accessible data, which results in out-dated FMs, i.e., GPT-4~(gpt-4-0125-preview) model's knowledge cut-off date is around 2023 December, which may fail to adapt to up-to-date use cases. 

FL provides a solution by facilitating the fine-tuning of FMs with incremental data and adapting FM to the new data ~\cite{zhang2023fedcontinual}. 
Rather than setting up complex data collection pipelines and starting the training process from scratch when new data becomes available, FL can integrate newly collected data into the existing FMs. This continuous model improvement ensures the FM remains up-to-date, readily adapting to dynamic-changing scenarios.
This continuous model improvement ensures that the FMs remain up-to-date and easily adapt to the dynamic-changing real world scenarios.

\textbf{Response Delay and Service Down in FM.} 
One of the key advantages of leveraging FL for FM is the ability to provide instant response, 
thus providing better user experience. Traditional approaches 
run FMs on the central server with API access, which can introduce both latency and privacy issues due to network communication between the users and the central server
~\cite{li2023snapfusion}. FMs on the cloud may also suffer from service down due to a series of unexpected circumstances \cite{chatgpt-down}.
On the contrary, 
FMs fully or partially deployed in FL participants significantly reduce inference latency, out of service and potential privacy problems. 

Overall, FM was motivated to be empowered by FL to overcome limitations related to data availability, hardware resources, FM monopoly, continuous data growth, 
response delay and service down. These advancements pave the way for the widespread adoption and further development of the FM, enabling its application across various domains.

\subsection{Challenges of Empowering FM with FL}

\begin{figure*}[t!]
\centering
\begin{subfigure}[t]{0.26\textwidth}
    \includegraphics[width=\textwidth]{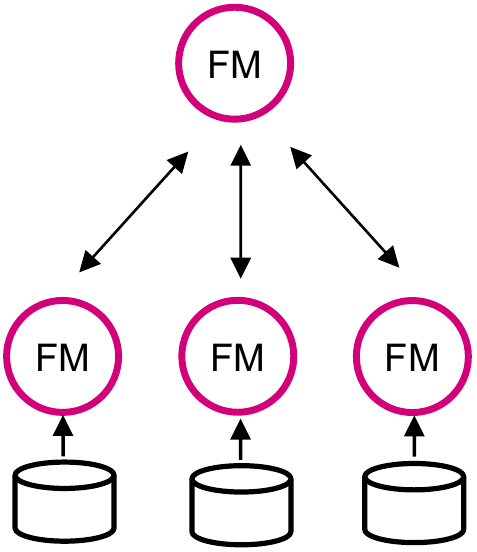}
    \caption{Train or fine-tune FM}
    \label{fig:fl-fm-full}
\end{subfigure}
\hfill
\begin{subfigure}[t]{0.26\textwidth}
    \includegraphics[width=\textwidth]{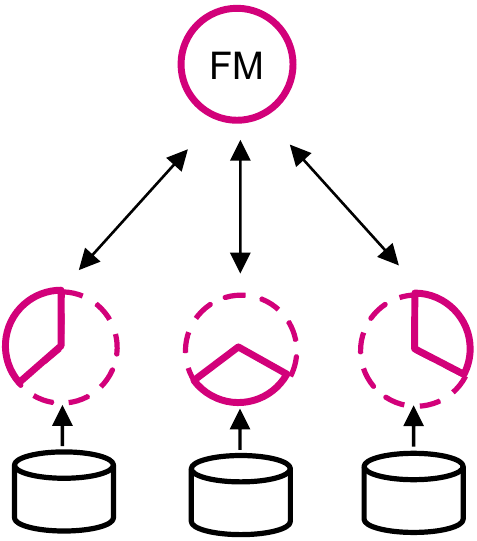}
    \caption{Parallel FM training}
    \label{fig:fl-fm-parallel}
\end{subfigure}
\hfill
\begin{subfigure}[t]{0.34\textwidth}
    \includegraphics[width=\textwidth]{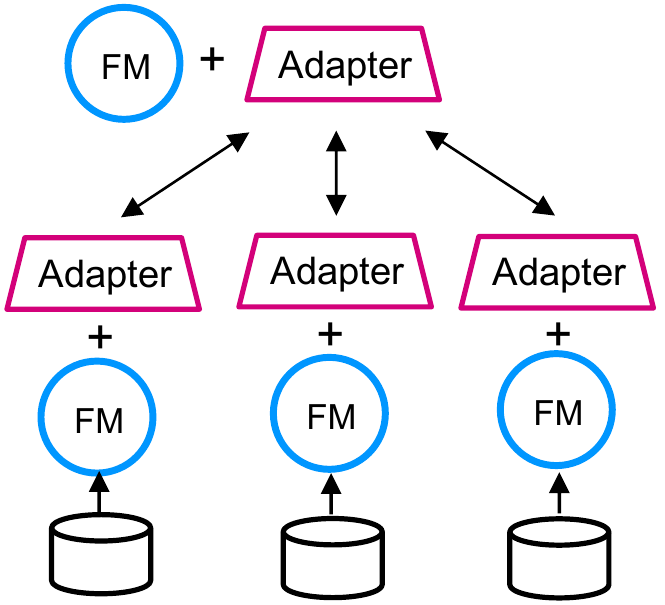}
    \caption{Fine-tune FM with adapter}
    \label{fig:fl-fm-adapter}
\end{subfigure}
\hfill
\begin{subfigure}[t]{0.1\textwidth}
    \includegraphics[width=\textwidth]{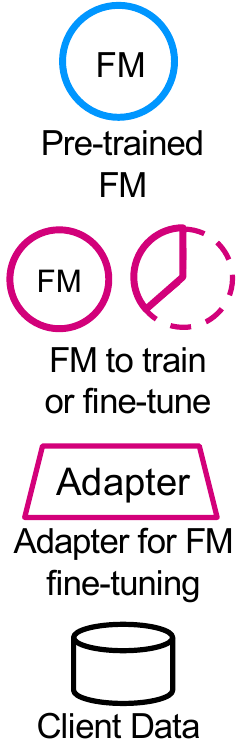}
\end{subfigure}
\caption{
Illustration of possible approaches for empowering Foundation Model (FM) with  Federated Learning (FL): (a) FL can be utilized to train FM from scratch or fine-tune FM using private client data. (b) Another approach is to utilize FL to train FM from scratch with model parallelisms (splitting the model into parts to train in parallel) using private client data. (c) FL can also be employed to 
allow clients to conduct parameter-efficient fine-tuning for their local FMs by training an additional adapter or using prompt tuning methods. These approaches illustrate the potential of empowering FM with FL.}
\label{fig:fl-for-fm}
\end{figure*}

This section explores challenges in empowering FM with FL. \cref{fig:fl-for-fm} illustrates potential approaches 
of using FL for FM. Both cross-silo and cross-device FL present opportunities to empower FMs, but they also face shared or unique challenges.

\textbf{Large FMs Incur High Memory, Communication, and Computation in FL.} 
Empowering FM with FL poses significant challenges due to substantial size of most FMs, resulting in high memory, communication, and computation requirements. 
For instance, 
open-sourced models like CLIP~\cite{radford2021clip} with 307M 
ViT-large image encoder occupies around 1.7 GB; Stable Diffusion XL~\cite{rombach2022high} has over 6.9 GB memory footprint, and the smallest LLaMA 3~\cite{dubey2024llama} has over 8 billion parameters. 
\cref{fig:fl-fm-full} depicts the typical approach in FL, where 
each client trains or fine-tunes its local FM and transmits FM updates to the global FM server. However, the large size of FM introduces multiple challenges, which are 
categorized into two main aspects:

(1) \emph{Difficulty in hosting FMs on FL clients}. Hosting large FMs is resource-intensive and could be problematic in FL, especially for clients with limited storage and computational capabilities \cite{lim2020federated,zhuang2022smart}. In addition to storing the default model size, extra high-speed memory is often required to store inputs and intermediate activation for local inference and finetuning. Participants in cross-silo FL must provide substantial infrastructure support for training or fine-tuning these large models. This challenge amplifies in cross-device FL, where client devices, such as smartphones or IoT gadgets, have significantly less memory and computational power compared to more powerful participants in cross-silo FL.

(2) \emph{Huge communication cost associated with sharing FM in FL}. Communication overhead in FL involves transmitting model parameters or updates between clients and the server \cite{shahid2021communication}. Transmitting extensive FM parameters over a network, especially with limited or unreliable bandwidth, can be time-consuming and resource-intensive. 
It can also lead to incomplete updates if the transmission is interrupted. 
Another subtle yet important issue is the increased risk of model and data staleness in FL. As multiple distributed clients are training the model, slow transmission of FMs can cause some clients to be training the outdated version of the model by the time they are received, which affects the convergence and overall performance.

\textbf{Challenges of Data Privacy, Copyright, and Model Security in FL for FM}. Empowering FM with FL introduces critical challenges in data privacy, copyright and model security. In both cross-silo and cross-device FL, 
malicious participants may exist and cause threats such as membership inference and model inversion attacks compromising privacy by extracting sensitive information from FMs like Stable Diffusion \cite{carlini2023extracting,rombach2022stablediffusion}. While techniques like deduplication, adopted in DALL-E 2 \cite{dall-e-2}, can help mitigate memorization \cite{carlini2023extracting} and the associated privacy risks, their application in FL is complex due to the distributed nature of data across multiple participants. 

Copyright is another important aspect to consider in order to ensure legal compliance and ethical usage of data. It is crucial to establish mechanisms that verify the ownership and rights of the contributed data by all participants. In the event of problematic data that violates IP or copyright, efficient and effective methods must be implemented to unlearn the potential impact on the FM.

Model security 
shall be considered when FL is applied to FM. 
In addition to the traditional security vulnerabilities from the participating clients through data or model poisoning attacks and Byzantine attacks \cite{fang2020local, lyu2022privacy}, new security threats are emerging targeting FM. For instance, recent research~\cite{ye2024emerging} highlighted the vulnerability of federated instruction tuning of large language models to safety attacks, where malicious clients can manipulate the model's behavior by injecting harmful instructions. ~\cite{zhao2024defending} pointed out parameter-efficient fine-tuning (PEFT), a trick commonly used by FL for FM, is more susceptible to weight-poisoning backdoor attacks compared to the full-parameter fine-tuning method.
The fusion of knowledge from multiple clients into the FM makes it crucial to ensure the integrity, confidentiality, and authenticity of the model. Breaches in security could lead to compromised models and result in severe consequences, as FMs will be utilized for 
various downstream applications, potentially impacting quality, user experiences, even causing harms.

\textbf{Complex Incentive Mechanisms for Collaboration in FL for FM.} Designing effective incentive mechanisms for collaboration in FL presents an additional challenge. 
Building an ecosystem to train the FM requires careful consideration of incentivizing entities with varying data and computing power ~\cite{lyu2020towards,lyu2020democratise}. Given the FM's significant power for diverse downstream tasks, defining model ownership 
among participants is crucial. 
Moreover, when the FM generates profit through APIs or other means, allocating profits and costs fairly among stakeholders becomes a critical aspect. While existing incentive mechanisms in conventional FL with smaller models provide a good starting point \cite{lyu2020collaborative,sim2020collaborative}, adapting 
existing methods 
in the context of FM may raise new challenges.

\subsection{Opportunities and Future Directions}

In this section, we present the opportunities and future directions of FL for FM.

\textbf{Integrating FL into the Lifespan of FM.} 
FL can empower the entire lifespan of FMs, encompassing pre-training, fine-tuning, and inference. 

\emph{Pre-training:} Firstly, FL offers the opportunity to leverage large-scale and decentralized datasets from multiple participants during the pre-training phase. This collaborative learning process leads to the development of more 
generalized FMs that can effectively handle a wide range of tasks and domains.

\emph{Fine-tuning:} Secondly, FL plays a crucial role in the fine-tuning of FMs by incorporating incremental data from participants. This iterative process ensures that the FMs remain up-to-date and adaptable. Recent advancements, such as GPT-JT \cite{gpt-jt}, have demonstrated the potential of obtaining powerful FMs from decentralized participants. Notably, GPT-JT, with only 6 billion parameters, outperforms many FMs with significantly larger sizes. Besides, we can use personalized FL techniques to obtain a personalized model for each client during the fine-tuning process.

\emph{Inference:} Thirdly, FL facilitates collaborative distributed model inference through both the central FM on the server and personalized FMs distributed in clients. Prompting emerges as a new method to interact and make inferences with FMs. Users can employ prompts directed either to the central FM on the server, to specific personalized FMs on clients, or a combination of both. Recent work \cite{zhao2023lingualinked} shows the potential of distributed inference of large language models
(LLMs). Exploring optimal coordination strategies among FMs remains open for further research.

\textbf{Memory, Communication, and Computation Reduction.} Addressing the challenges associated with memory, communication, and computation in FL for FM opens up promising avenues for improvement. There are several techniques that can be explored in this regard.

(1) \textit{Efficient distributed learning algorithms} could be adopted to address the communication and computation challenges \cite{zhao2022reduce, wu2022communication}. 
Parallelism, such as model and pipeline parallelism, is a technique from distributed training \cite{dean2012large, huang2019gpipe}. Model parallelism partitions a model and distributes it across devices for parallel processing, while pipeline parallelism improves efficiency and scalability. \cref{fig:fl-fm-parallel} illustrates the idea of adopting parallelism in FL for FM, where participants train partial layers of a model using private data. Recent research explores decentralized FM training with pipeline parallelism \cite{yuan2022decentralized}. Future studies can explore 
efficient approaches and decentralized optimization for parallelisms in FL for FM.

(2) \textit{Parameter-efficient training methods} can adapt FMs to specific domains or tasks \cite{houlsby2019petl-adapter,hu2021lora}. These methods commonly freeze the FM parameters and fine-tune only a small adapter or train a Low-Rank Adaption~(LoRA) module. \cref{fig:fl-fm-adapter} depicts a possible approach for adopting the adapter. Recent works like FedCLIP \cite{lu2023fedclip}, FFM \cite{yu2023federated}, FedDAT~\cite{chen2024feddat}, FedAG~\cite{wang2024fedadapter} and FedLoRA~\cite{yi2023fedlora} adopt this method and achieve significant performance improvement. However, participants still require substantial computational resources to fit the FM and execute the fine-tuning. 

(3) \textit{Prompt tuning} has been shown to be as effective as model tuning \cite{lester2021prompttuning}. Instead of sharing model parameters of large FMs, prompt tuning offers a new communication-efficient sharing paradigm. This technique freezes the entire FM's parameters and fine-tunes additional lightweight tokens. Recent studies like FedPrompt~\cite{zhao2023fedprompt}, PROMPTFL~\cite{guo2022promptfl}, PFPT~\cite{weng2024probabilistic}, FedBBPT~\cite{lin2023efficient}, and FedBPT~\cite{sun2023fedbpt} showcased the potential of leveraging FL to enhance the quality and effectiveness of prompts in FM training. However, these studies either assume the presence of large FMs on user devices, which may not be feasible in resource-constrained cross-device settings, or face privacy concerns by relying on cloud FM APIs that require sending the sensitive data. 
Further research should consider the constraints of devices. 
Exploring methods to refine prompts in FL, leveraging knowledge from multiple participants, is promising to improve FM performance.

(4) \textit{Advanced model compression techniques} are vital for reducing the high memory, communication, and computational costs of
FMs~\cite{song2022resfed,huang2024fedmef}. Model compression can reduce the size of FMs without a significant loss in model performance, thereby making them more manageable for deployment in FL. Common mechanisms 
can be considered for adaptation to FMs. For example, through knowledge distillation \cite{hinton2015distilling}, a smaller model is trained to replicate the behavior of a larger one; quantization \cite{han2015compression,nagel2021white} reduces the numerical precision of the model's parameters, further reducing model size and computational requirements. Despite existing efforts in reducing the computation complexity of FM such as Linear Attention~\cite{koohpayegani2024sima}, Flash Attention~\cite{dao2022flashattention}, future research should aim to push the boundaries of these techniques, striving to optimize the efficiency of FMs to the greatest extent possible without compromising their ability to generate valuable insights.

Moreover, these techniques can be further combined. 
For instance, Offsite-Tuning \cite{xiao2023offsite} applies compression methods \cite{han2015compression,hinton2015distilling} and the concept of parameter-efficient tuning \cite{houlsby2019petl-adapter} to gain knowledge from a single private data owner.
Emerging applications in the context of FL and FMs have already been demonstrated \cite{kuang2024federatedscope, wu2024fedbiot}. Future research could further explore leveraging these combined approaches to optimize FMs within FL involving multiple participants.

\textbf{Designing FL System and Benchmark for FM.} Designing specialized FL systems and benchmarks is important to cater to the unique requirements and challenges of FM~\cite{wangfedmeki}. While existing FL systems have made significant progress in both academia and industry \cite{bonawitz2019towards,he2020fedml,huba2022papaya,beutel2020flower,liu2021fate, zhuang2022easyfl,xie2023federatedscope, zhuang2024coala}, they may not fully address the optimization of memory, communication, and computation specifically for FM. Platforms like FedML \cite{fedml-llm}, FATE \cite{fate-llm} and COALA~\cite{zhuang2024coala} are evolving to support FM, incubating research works such as FedAG~\cite{wang2024fedadapter}. Due to the complexity of system requirements and integration strategies, further research is needed. This involves considering scalability, efficiency, and privacy aspects in FM training and inference, and devising mechanisms to handle the large model sizes and resource constraints. Additionally, the development of comprehensive benchmarks tailored for FM can facilitate fair comparisons and evaluation of FL algorithms. These dedicated FL systems and benchmarks for FM can advance the state-of-the-art in FL, enabling more efficient and effective collaborative and privacy-preserving FM training.

\textbf{Improving FM with Decentralized Data in FL.} 
While FMs are typically trained on large-scale centralized datasets, there may exist domain gaps and limitations in their coverage of real-world data distributions \cite{hong2023mecta}. FL can help bridge these gaps by incorporating data from diverse sources, ensuring a more comprehensive representation of different scenarios and improving the generalization capabilities of FMs. Moreover, the increasing computational power of personal devices, like laptops and mobile phones, presents an opportunity to make FL more accessible for FM. 
For example, FL can leverage the enhanced computation for fine-tuning to continuously incorporate knowledge back into FMs. 
While training an FM from scratch using cross-device FL may currently be challenging due to memory and computation limitations, advancements in device capabilities and model size reduction pave the way for potential future developments.

\textbf{Exploring Incentive Mechanisms in FL for FM.}

Incentive mechanisms are studied in conventional FL with smaller models \cite{lyu2020collaborative}, the unique challenges from FMs require tailored solutions. Designing fair incentives that consider varying data and computing power across 
participants is essential for encouraging participation and collaboration. 
This includes addressing disparities in data contributions and computational resources, ensuring that incentives are aligned with the effort and value provided by each participant. Additionally, determining profit and cost allocation after deploying FMs is crucial. 
Fair distribution of rewards and benefits is necessary to maintain trust and encourage ongoing engagement. Mechanisms should be established to determine how profits generated from FMs are allocated among the participants. 
This ensures an equitable distribution of economic benefits and encourages continued participation and investment in FL for FM.

\textbf{Advancing Trustworthy FL for FM.} 
Trustworthy FL for FM requires addressing various aspects, including privacy, security, 
fairness, interpretability, 
IP, and more. To enhance the privacy and security of FL for FM, mechanisms need to be developed to detect and prevent attacks from malicious participants. 
Resolving biases in the training data and improving the interpretability of FMs are important for developing fair and accountable models. Additionally, protocols and frameworks should be established to address IP and copyright concerns when incorporating data from multiple clients in FL, while also considering the issue of model ownership. Moreover, 
how to address the memorization problem in FL for FM presents an important opportunity for further studies. Deduplication 
previously was applied in centralized settings by identifying similar images \cite{dall-e-2}. A distributed deduplication mechanism 
is needed in FL for FM. 

\begin{figure*}[t!]
\centering
\begin{subfigure}[t]{0.24\textwidth}
    \includegraphics[width=\textwidth]{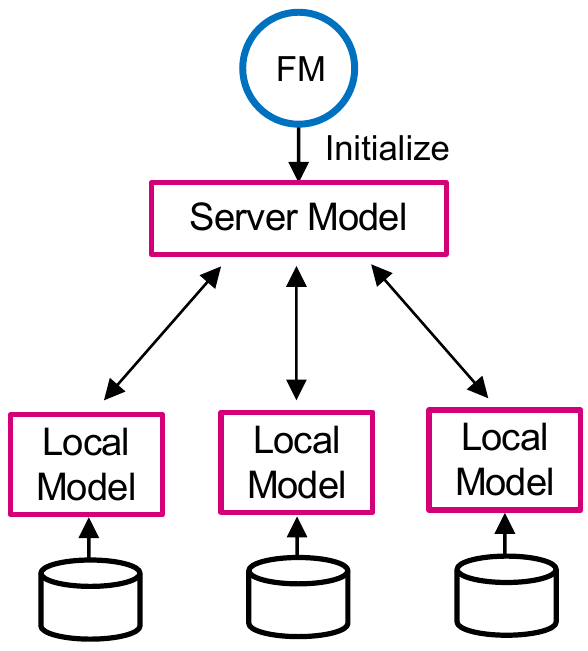}
    \caption{FM for better FL initialization}
    \label{fig:fm-fl-initialization}
\end{subfigure}
\hfill
\begin{subfigure}[t]{0.62\textwidth}
    \includegraphics[width=\textwidth]{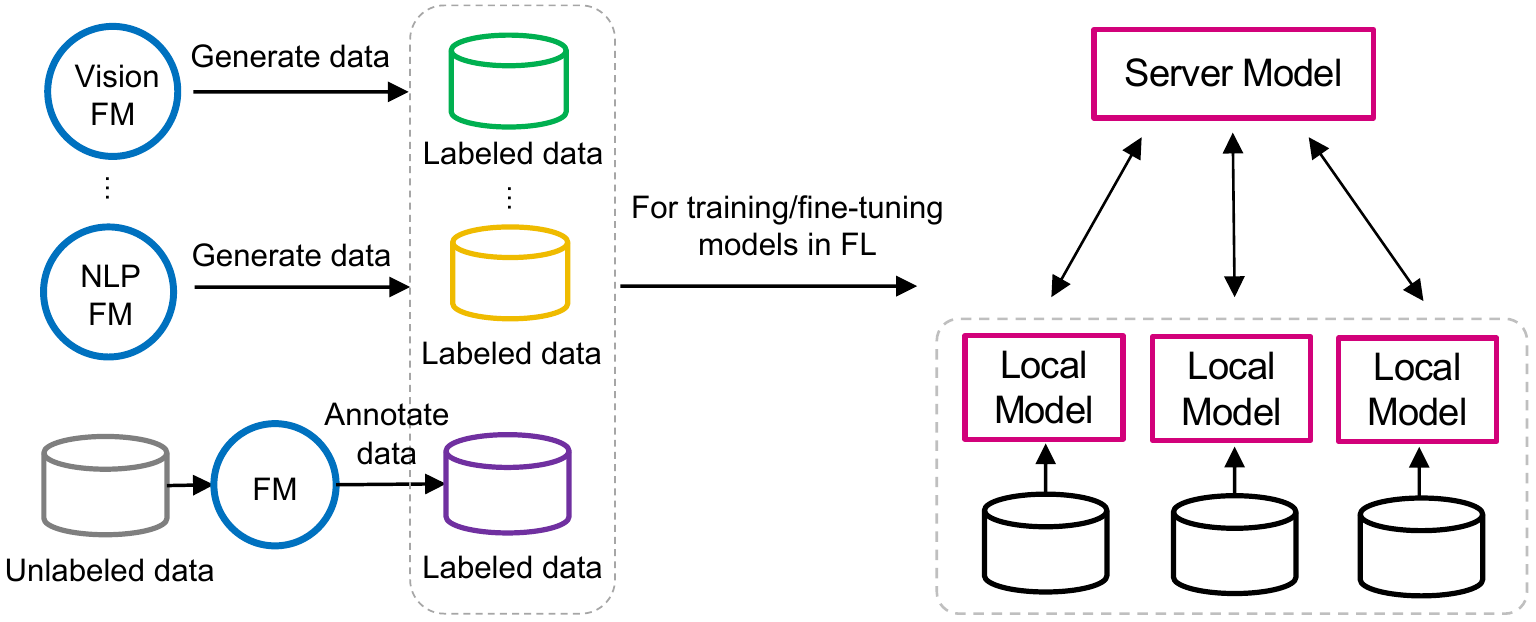}
    \caption{FM data generation and annotation for FL}
    \label{fig:fm-fl-data-synthesis}
\end{subfigure}
\hfill
\begin{subfigure}[t]{0.09\textwidth}
    \includegraphics[width=\textwidth]{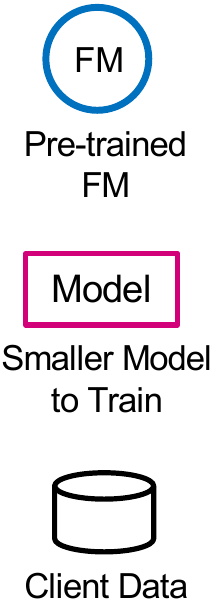}
\end{subfigure}
\caption{
Illustration of possible approaches of using Foundation Model (FM) to address suboptimal performance of Federated Learning (FL): (a) FM or FM after compression or distillation can be used as a more powerful initialization for FL. (b) One or multiple FMs can be used to synthesize or annotate data, each of which then can be used independently or combined for training models on the server or FL clients.}
\label{fig:fm-for-fl} 
\end{figure*}

\section{Foundation Models for Federated Learning}

In addition to the advantages that Federated Learning (FL) brings to the Foundation Model (FM), FM can also contribute to FL. 
In this section, we delve into the motivation behind FM for FL, examine the challenges that arise in this context, and explore the potential opportunities in FM for FL.

\subsection{Motivations of FM for FL}

\textbf{Suboptimal Performance and Privacy Dilemma in FL Needs Help from FM.}
FL often suffers from performance degradation, especially under non-iid and biased settings, where FL clients have limited or imbalanced data, such as federated few-shot learning contexts~\cite{shome2021fedaffect}. This data scarcity cannot fully representing the diversity of the data distribution, thus impacting model performance~\cite{gao2022survey}. Furthermore, several studies highlight that sensitive data information can be recovered from model updates in FL \cite{lyu2022privacy,chen2022practical}.
These issues are amplified in sectors like healthcare or finance, where data is highly sensitive or difficult to collect due to data privacy regulations or the sensitive nature of the data itself~\cite{zhang2022dense}. 
FMs can help improve the performance of FL through the following ways: 

(1) FMs can serve as a good starting point or serve as a good teacher for FL, as shown in \cref{fig:fm-fl-initialization}. Clients can utilize the well initialized models with the inherited knowledge from FMs and directly conduct fine-tuning on their local data without training from scratch~\cite{chen2023importance,nguyen2022begin,tan2022federated}. By transferring knowledge from FMs to the models in FL system, it accelerates convergence, mitigates the suboptimal performance issues caused by limited or imbalanced data, and reduces communication costs by reducing numerous rounds of communication. 

(2) Generative FMs, pre-trained on extensive datasets and refined through fine-tuning, have better robustness and generalizability. The generated synthetic data can enrich the diversity and reflect the complexity of real-world scenarios. \cref{fig:fm-fl-data-synthesis} illustrates a feasible approach for using FM to generate synthetic data for FL. In cross-silo FL, clients with computational resources can use FMs to synthesize data for local model training; while in cross-device FL, the server can generate synthetic data using FMs. This approach offers several key benefits: (1) Synthetic data can alleviate the data scarcity for clients with limited training data, or serve as the data for knowledge distillation in FL~\cite{zhang2022dense}; (2) Synthetic data introduces diversity into training, especially for scenarios not covered in the original data ~\cite{de2022mitigating}, which can help mitigate overfitting and enhance model generalization; (3) Synthetic data mitigates privacy concerns by supplementing or replacing sensitive training data~\cite{dong2022privacy,dockhorn2022differentially}, 
thus 
avoiding privacy leakage in FL. 
For language tasks, \cite{huang2022large, wang2022self} showcased the potential to further improve LLM with generated data. For vision tasks, 
~\cite{zhang2023diffusionengine}, ~\cite{xin2024dart} and ~\cite{li2024simple} used diffusion model to scale up the data volume and improve object detection models. In a similar way, generated data can be used to improve various tasks in FL. GPT-FL~\cite{zhang2023gpt} leveraged generated synthetic data to get a better model initialization for FL participants. \cite{wang2024fedadapter} used stable diffusion with style information to conquer the data heterogeneity in FL. These evidences demonstrate strong potential to outperform state-of-the-art FL methods thanks to the general insights from FMs.

(3) FMs can also serve as effective annotators, providing pseudo-labels 
for FL participants, as illustrated in \cref{fig:fm-fl-data-synthesis}. Recently, FMs such as GPT-4-o~\cite{achiam2023gpt}, SAM~\cite{kirillov2023sam}, and GroundingDino~\cite{liu2025grounding} have been adopted~\cite{roboflow2025} to support annotation tasks for users who cannot afford the expense of manual labeling, particularly for challenging tasks such as segmentation and object detection. Leveraging the annotation capabilities of FMs, studies like~\cite{yang2024boosting} and~\cite{yuan2024unsamflow} have demonstrated success in improving specific tasks, such as edge detection and optical flow. Furthermore, multiple FMs can be employed in combination to generate multi-modal annotations~\cite{xiao2024omnigen, xiao2024florence}.
Recently, DART~\cite{xin2024dart} highlighted the use of GPT-4-o~\cite{achiam2023gpt} as a pseudo-label evaluator, further enhancing the annotation quality of FMs. Moreover, when combined with real data in semi-supervised FL scheme such as~\cite{jeong2020federated, lin2021semifed}, performance degradation of noisy pseudo-label can be further minimized. By providing high-quality pseudo-labels, FMs significantly lower the barrier to entry for FL.

\textbf{New Sharing Paradigm Empowered by FM.} In contrast to the conventional approach of sharing high-dimensional model parameters in traditional FL, FMs pave the way for a novel sharing paradigm by incorporating the concept of prompt tuning.
PROMPTFL~\cite{guo2022promptfl} demonstrated how the capabilities of FMs can be harnessed to simultaneously facilitate efficient global aggregation and local training on sparse data. This is achieved by allowing FL participants to focus on training prompts rather than a shared model.
Following a similar trajectory, Zhao et.al. \cite{zhao2022reduce} introduced FedPrompt, an innovative prompt tuning method tailored for FL. This approach involves keeping pre-trained language models (PLMs) fixed and focusing solely on the aggregation and tuning of 'soft' prompts, with the objective of reducing communication overhead. This paradigm shift brought by FM, pivoting from shared model parameters to shared prompts, sets a new precedent in the field of FL.

\textbf{Multi-task and Multi-modality Capability Empowered by FM.} 
Most FL models are limited to a single task, domain, or modality, posing challenges for real-world scenarios, such as autonomous driving, which requires handling multiple tasks concurrently~\cite{zhuang2023mas,zhuang2024fedwon}. FMs present a promising solution to this constraint. FMs, especially those predominantly equipped with multi-task and multi-modality capabilities~\cite{wang2022image,xiao2024florence,zhang2023next, achiam2023gpt}, offer a compelling avenue for addressing diverse task requirements. Training a multi-task and multi-modal model via FL from scratch is challenging \cite{zhuang2023mas}; however, multi-task and multi-modal FMs can be compressed or distilled to a more powerful multi-task multi-modal initialization for FL like \cref{fig:fm-fl-initialization}. Also, they can facilitate multi-modal data and annotation generation for diverse FL tasks like \cref{fig:fm-fl-data-synthesis}.
By adeptly incorporating FMs 
into FL system, FL models can gain the capacity to handle multiple tasks and modalities seamlessly.

\subsection{Challenges of Empowering FL with FM}
\label{sec:fm-fl-challenges}

\textbf{Problematic Synthetic Data Generated by Generative FM.}
Although generative FMs can generate abundant synthetic data, the quality of the generated data cannot be guaranteed. Firstly, the data sources can be problematic. Large-scale data scraped from the Internet may be of low quality, containing bias, misinformation, and toxicity. FMs trained on such data without mitigation strategies may inherit harmful stereotypes, biases, and errors, which can be further exacerbated in the generated data. For example, the LAION dataset~\cite{schuhmann2022laion} used for training Stable Diffusion may unintentionally include bias and toxic data, potentially inherited by the synthesized images. Thus, it is crucial to sift out clean subsets when using FMs to generate data for FL training~\cite{zeng2022sift}. Secondly, the generated data could be problematic, especially when the synthetic data does not align with clients' data distribution. It may introduce bias and noise into FL training~\cite{mohri2019agnostic}. For example, over- or under-representation of certain classes or features in synthetic data can skew the resulting FL model, leading to poor generalization and biased predictions. For LLM, it has been identified that the overuse of synthetic data may cause issues such as model collapse~\cite{shumailov2024ai}, leading to compromised performance.

\textbf{Law Compliance and Responsible Technology Usage of FM for FL.} 
Despite the promising potential of FMs for FL, it also poses new challenges related to privacy, IP rights, and the responsible use of FMs. 
These challenges arise from the fact that many FMs are trained on voluminous data, often acquired from the internet without adherence to the proper legal and ethical protocols. 

(1) \emph{Privacy and IP law violation.} One significant challenge is the formidable memorization capability of FMs~\cite{carlini2023extracting,somepalli2022diffusion,somepalli2023understanding,wen2024detecting,ren2024unveiling}. These models have the capacity to reproduce data directly from the training sets, a feature that raises serious privacy concerns~\cite{stable-diffusion-litigation-url}. If the synthetic data generated by a FM is too closely aligned with the original training data, the risk of revealing sensitive or proprietary information and copyright infringement increases substantially. 
Therefore, analyzing the origin of synthesized data becomes necessary~\cite{wang2023alteration}. This is not a trivial task as it involves tracing the lineage of synthetic data back to its original sources and verifying its uniqueness. Such processes require substantial computational resources and sophisticated algorithms, posing an additional challenge in resource-constrained FL scenarios. 

(2) \emph{Misuse of technology.} 
The potential misuse of FMs within the context of FL presents a significant and growing concern. 
As FMs become more powerful and widely integrated into decentralized systems, the risks associated with their exploitation for malicious purposes — such as generating harmful content, spreading misinformation, or breaching privacy — escalate dramatically~\cite{chen2023pathway}. 
Safeguarding against such misuse requires a robust framework of ethical guidelines and regulations, which adds to the complexity of empowering FL with FMs.

\textbf{Unique Challenges in Adopting FM for FL}.
Current research on empowering FL with FMs is still in its nascent stage. With a considerable gap between the current state of research and practical deployment. We highlight five challenges to tackle in real-world scenarios: (1) \emph{Label scarcity.} Practical scenarios often involve scarce or absent labeled data~\cite{zhuang2021fedureid,li2022mocosfl,hong2022outsourcing,liu2022optimizing}. It requires further exploration of unsupervised or semi-supervised methods in using FMs for FL. (2) \emph{Continuously changing environment.} Real-world data arrives as a continuous stream ~\cite{kirkpatrick2017overcoming,zhang2023fedcontinual}, but most 
research currently centers around using static FMs for FL~\cite{zhang2023gpt}, which might not effectively capture the evolving data. The use of dynamic FMs, which can be updated and adapted over time, could be a potential solution. It needs further study on how FMs can benefit FL 
in the continuously changing environment. (3) \emph{Domain gap.} The domain gap between 
data used by different models or systems is another challenge often understated \cite{glorot2011domain,long2015learning,zhuang2021fedfr,hong2023mecta,tan2023heterogeneity,fan2023federated}. In practice, there can be significant differences between training data used by FMs and the data encountered by FL systems. How to choose the most suitable FMs for different FL systems pose practical challenges.
(4) \emph{White-box vs black-box access to FMs.} 
Due to the huge profit market of FMs, many FMs are only available in a black-box manner via API~\cite{hecater,he2022protecting}.
In the absence of white-box access to FMs, it becomes more challenging to bring the benefits of the most powerful FMs to FL.

In conclusion, while the preliminary research into the use of FMs for FL is promising, there is a clear gap between research and practical deployment.

\subsection{Opportunities and Future Directions}

In this section, we discuss the opportunities and potential future directions of empowering FL with FM.

\textbf{Legal Consideration and Responsible Usage of FM for FL.}
As discussed in Section \ref{sec:fm-fl-challenges}, it is crucial to ensure the use of FMs does not infringe upon privacy and IP laws, especially when using synthetic data generated by FMs. We further provide several possible ways to address this challenge: 

(1) \emph{License check and thorough assessment of FMs}. It is recommended to use more secure FMs with CC by 4.0 or MIT licenses, as well as those that are more transparent and responsible, rather than less transparent ones with vague licenses. FMs that open-source their training data, code, and checkpoints can be assessed more thoroughly before applied to FL. Conversely, the hidden risks of close-sourced FMs are unknown, which may raise serious consequences when applied to FL. Therefore, before using FMs for FL, there is a high necessity to conduct comprehensive assessments regarding to its license, open-source status, and responsibilities, including training data copyright, privacy, bias, and model security.

(2) \emph{Verification of synthetic data}. Before adopted by FL, it is worthwhile to validate the synthetic data generated by generative FMs, which calls for advanced methods. For example, statistical or machine learning methods can be designed to detect similarities between synthetic and real-world data~\cite{wang2023did,wang2023alteration,wang2024trace}. It is essential that these methods can accurately determine if the synthetic data is sufficiently different from any real-world data to avoid privacy and IP infringements. 
Besides, designing more faithful metrics for the privacy assessment is also worthwhile to explore~\cite{sun2023privacy}. 

(3) \emph{Privacy-preserving data synthesis}. Exploring privacy-preserving techniques tailored for FMs could be an important direction. This could involve research into methods that allow FMs to generate synthetic data or distill their knowledge while inherently preserving privacy. 
Privacy-preserving techniques in model training and inference should also be explored to safeguard against the exposure of sensitive information.

(4) \emph{Interdisciplinary collaboration}. Addressing these challenges would require a multi-disciplinary approach, involving not only advances from machine learning but also contributions from law, ethics, and social sciences. Interdisciplinary collaboration is essential to ensure the development of robust and ethically grounded methods in this evolving research direction.

\textbf{Enhancing Robustness of FM for FL.}
The application of FMs to FL introduces challenges that may impact the robustness of the FL-trained models. The following are four potential opportunities to achieve better robustness:

(1) \emph{Robustness against bias.} FMs trained on data skewed towards a specific demographic may introduce biases in synthetic data or distilled knowledge. This could lead to fairness issues in the FL model, where the model exhibits superior performance for the favored demographic and inferior performance for others. Future research should address this and prevent the amplification of existing biases by developing techniques for fair data augmentation, fair knowledge distillation, and methods to assess and minimize biases.

(2) \emph{Robustness against out-of-domain generation.} Ensuring that synthetic data accurately represents the underlying client data distribution is critical for improving model performance in FL. If the synthetic data generated by FMs could not accurately represent the underlying data distribution of FL clients, it could lead to model performance degradation. Future research should focus on enhancing the quality and diversity of synthetic data produced by FMs, to better align with underlying data distributions, domains, and modalities in the FL context.

(3) \emph{Robustness against poisoned models.} Using a compromised FM for FL can have detrimental effects. For example, a diffusion model with a concealed "backdoor" could initiate harmful activities when it detects a specific trigger pattern ~\cite{chou2022backdoor,sun2023defending}. This Trojan effect could inflict severe harm on downstream applications that rely on the compromised diffusion model. Future research should investigate how the robustness of FM impacts FL when FM is used for FL. 

(4) \emph{Robustness against byzantine attackers and adversarial examples.}
Another two important robustness concepts in FL are robustness against byzantine clients~\cite{chen2022calfat,liu2023byzantine} and robustness against adversarial examples ~\cite{zhang2023delving, wang2023robustness}. The integration of FM into FL introduces new challenges to the robustness of FL, such as the adversarial prompts introduced by byzantine attackers. Ensuring client loyalty is crucial to prevent compromise of the entire FL system. Similarly, it is also crucial to safeguard FMs against these issues because the FL system would inherit the vulnerability if FMs are vulnerable to adversarial examples.

\begin{figure}[t!]
    \centering
        \includegraphics[width=0.48\textwidth]{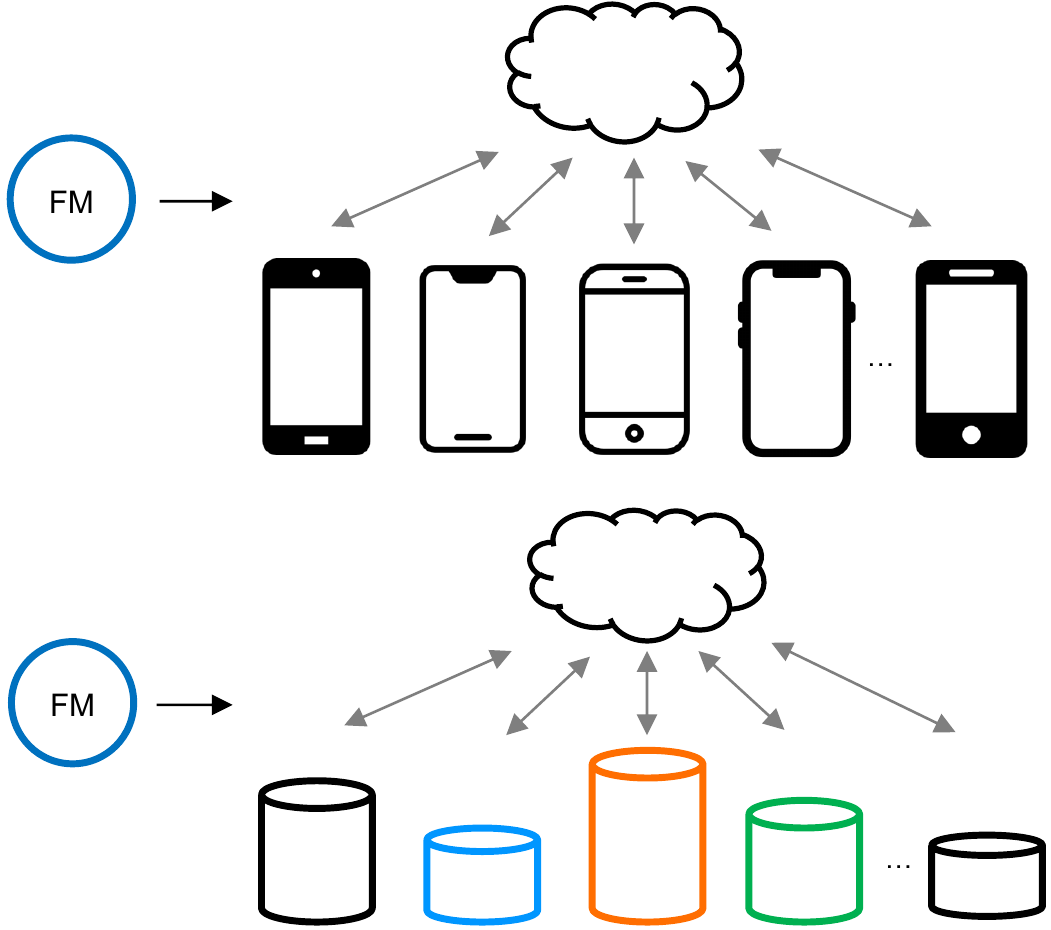}
        \caption{Illustration of the opportunity of personalizing FM for FL clients with system heterogeneity (top) and data heterogeneity (bottom).}
        \vspace{-3ex}
    \label{fig:fm-fl-personalization} 
\end{figure}

\textbf{Personalization of FM for FL.} 
Exploring personalized FMs for diverse FL scenarios 
matters a lot for real-world applications~\cite{tan2022towards}, especially under system heterogeneity where clients vary significantly in memory and computation power. Given the remarkable zero-shot or few-shot capabilities of FMs, personalizing these models holds potential benefits for FL clients to deal with data heterogeneity and unseen domains. \cref{fig:fm-fl-personalization} illustrates the two key areas to investigate:

(1) \emph{Tailoring FMs to FL clients with heterogeneous system}. One future direction lies in developing techniques that allow FMs to better cater to 
different memory and computational constraints of FL clients~\cite{wangflora}. This could involve methods for adapting FMs to specific applications, or algorithms dynamically adjusting model complexity based on clients' memory and computational resources. Future work could also explore strategies for selectively transferring knowledge from FMs to local models to meet each client's specific needs.

(2) 
\emph{Tailoring FMs to FL clients with heterogeneous data}. Data heterogeneity is a critical aspect in real-world FL applications~\cite{tan2023heterogeneity}. 
Personalization of FMs in FL offers a powerful solution for adapting FMs to individual clients, allowing for fine-tuning based on their unique data distributions. For example, personalized FMs can be leveraged to generate personalized synthetic data for FL clients. FMs, with their ability to learn rich and generalized representations, have the potential to create high-quality synthetic data tailored to the specific characteristics of each client’s local data distribution. This synthetic data could not only help address challenges for clients with imbalanced or limited data but also make the model more resilient to domain shifts.
Future research could explore how to further optimize FM personalization in FL to better address data heterogeneity in real-world FL applications.

\textbf{Advanced Knowledge Distillation Techniques of FM for FL.} 
FMs with extensive pretrained knowledge have the potential to enhance FL models through knowledge distillation (KD). Traditional KD methods may be suboptimal in FL due to communication constraints and data heterogeneity. Future research could explore advanced KD techniques tailored to leverage FMs in FL. Additionally, three directions are worth further exploration:

(1) \emph{Low-cost KD  from FMs to FL}. 
Incorporating FMs into FL presents significant challenges, particularly related to the high computational costs associated with KD. KD is essential for transferring the knowledge from large, pre-trained FMs to smaller, more efficient models suitable for FL. However, the computational expense of distilling knowledge from these large models can be prohibitive, especially in large-scale deployments. To address this, it is crucial to develop efficient distillation techniques that allow for the extraction of valuable information from FMs while minimizing resource consumption. These low-cost KD methods could enable the widespread use of FMs in FL, making it possible for clients with limited computational power to benefit from the rich representations learned by large FMs, without incurring high training or communication costs.

(2) \emph{Fair and privacy-preserving knowledge transfer from FMs to FL clients}. Future work should not only prioritize performance but also consider the fairness and privacy implications of knowledge transfer process from FMs to FL clients. 
Developing techniques that balance effective knowledge transfer with fairness and privacy preservation will be the key to the ethical and responsible deployment of FMs in FL systems, ensuring that all participants benefit from the shared learning process while maintaining data privacy.

(3) \emph{Black-box KD from FMs to FL}. The goal of black-box KD is to transfer the knowledge from a pre-trained FM, referred to as the teacher model, to many models in FL clients, known as the student models, without direct access to the internal parameters or architectural details of the FM. Several techniques can be employed to achieve this goal. One common method is to utilize the outputs of teacher model as pseudo-labels for the training data in FL. By aligning the predictions of the student model with the pseudo-labels provided by the teacher model, the student model learns to mimic the behavior and knowledge of the teacher model. This process typically involves iterative steps of training the student model, comparing its predictions with the teacher model's outputs, and adjusting the student model's parameters to minimize the discrepancy between the two. 
However, this process also typically involves a public dataset in the same domain for alignment, which might be not feasible in many sensitive scenarios.
It is appealing to investigate whether data-free black-box KD techniques~\cite{zhang2023ideal} can be utilized to transfer the valuable knowledge from black-box FMs to FL clients.

\textbf{Real-world Applications of FM to FL.} 
Real-world scenarios present a myriad of challenges when implementing FMs into FL, such as managing unlabeled data, handling continuous data streams, addressing domain discrepancies, 
and navigating substantial heterogeneity in both data and system. 
Furthermore, integrating FMs into FL presents an avenue to elevate real-world applications in various ways as follows.

(1) \emph{How to effectively utilize multiple FMs for FL?}
Recent advancements, particularly with large language models (LLMs), have showcased remarkable capabilities in language understanding, generation, interaction, and reasoning. Shen et al.~\cite{shen2023hugginggpt} proposed HuggingGPT, which leveraged LLMs as controllers to effectively manage diverse AI models in addressing intricate tasks. The growing availability of FMs within open-source communities presents an opportunity for the seamless coordination among multiple FMs for real-world FL applications.

(2) \emph{How FMs can accelerate FL deployment?} Deploying FL in real-world applications demands substantial effort, such as data preprocessing, model development, and communication optimization. FMs possess inherent zero-shot capabilities across diverse data types and excel in transfer learning. Leveraging FMs as initial models or employing them to generate synthetic data could significantly expedite the landing of FL. This acceleration results in faster convergence rates, 
better performance within FL implementations, ultimately expediting the integration of FL models into real-world applications.

(3) \emph{How to leverage FMs for FL in specific industries?} Further exploration of FMs in FL entails investigating their adaptability within specific industry verticals. For instance, evaluating the effectiveness of FMs in healthcare for predictive analytics or exploring their role in enhancing privacy-preserving mechanisms within FL setups. Moreover, examining FMs' potential in facilitating cross-modal learning, where information from different data modalities is integrated, can unlock new possibilities for more comprehensive and diverse FL applications.

Moving forward, it is important to have more practical, application-oriented research. To bridge the gap between research and commercial deployment, future studies should not solely concentrate on the theoretical aspects of utilizing FMs in FL. The constraints and requisites of real-world FL applications should also be considered to ensure the practical viability of these efforts.

\section{Conclusion}

This work studies the synergistic relationship between Federated Models (FM) and Federated Learning (FL). We discuss the motivations, challenges, and future directions for empowering each with the strengths of the other. 
On the one hand, FL mitigates challenges faced in FM development by expanding data availability, enabling computation sharing, allowing collaborative FM development, keeping FM always updated and avoiding service delay and down. On the other hand, FM contributes to FL by serving as a robust training starting point, generating synthetic data to enhance performance, providing new sharing paradigm and new capability. This work also highlights the challenges and opportunities for future research. We hope that this paper provides a good foundation and strongly advocates for continued research efforts to drive advancements in both FM and FL.

\ifCLASSOPTIONcaptionsoff
  \newpage
\fi



\bibliographystyle{IEEEtran}
\bibliography{IEEEabrv,references}
\end{document}